\documentclass[11pt]{article}

\usepackage[T1]{fontenc}
\usepackage[utf8]{inputenc}
\usepackage{amsmath}
\usepackage{amssymb}
\usepackage{booktabs}
\usepackage{microtype}
\usepackage[margin=1in]{geometry}
\usepackage{xcolor}
\usepackage{hyperref}

\hypersetup{
  colorlinks=true,
  linkcolor=blue!55!black,
  citecolor=blue!55!black,
  urlcolor=blue!55!black
}

\newcommand{\R}{\mathbb{R}}
\newcommand{\E}{\mathbb{E}}
\newcommand{\norm}[1]{\left\lVert #1 \right\rVert_2}
\newcommand{\inner}[2]{\left\langle #1, #2 \right\rangle}
\newcommand{\code}[1]{\texttt{#1}}

\title{Clark Hash: Stateless Sparse Johnson-Lindenstrauss Quantization for Neural Embeddings}
\author{Clark Labs Inc. \and Autoresearch \and Stanislav Kirdey}
\date{May 2026}

\begin{document}
\maketitle

\begin{abstract}
Clark Hash is a small method for storing neural embeddings in less space. It
normalizes each database vector, applies a deterministic sparse signed
Johnson-Lindenstrauss projection, clips the result, and stores a fixed-width
scalar-quantized code. Queries stay in floating point and are scored against the
stored sketches. In the default 384-dimensional sentence-embedding setting,
Clark Hash stores a cosine-search vector in 48 bytes instead of 1536 bytes for
dense \code{f32} storage. This is 32x smaller. The method does not need a
training pass, learned codebooks, rotations, or corpus statistics before new
vectors can be stored. We describe the codec, the Rust implementation, and a
multilingual sentence-similarity evaluation on 9,304 labeled pairs from 29
subsets. With a multilingual MiniLM encoder, the 48-byte sketches reached 0.910
and 0.946 macro Pearson correlation with dense cosine scores on STS17 and
STS22. Clark Hash is not a new Johnson-Lindenstrauss theorem and it is not a
replacement for approximate nearest-neighbor indexes. It is a simple stateless
codec for compact embedding storage.
\end{abstract}

\section{Introduction}

Embedding systems often store many dense floating-point vectors. A single
384-dimensional \code{f32} sentence embedding uses 1536 bytes before index
overhead. That can become expensive in memory, local cache, disk, or network
transfer.

Many compression methods learn centroids, rotations, codebooks, or calibration
parameters from a corpus. These methods can work very well. They are less
convenient when vectors arrive one at a time and must be stored before a
training set is available.

Clark Hash is meant for this online case. Each database vector is encoded on its
own using a deterministic seed. Query vectors stay in floating point. The query
is scored against bit-packed database sketches. This paper documents the method
as implemented in the \code{clark-hash} Rust package. The implementation keeps
the original internal API name \code{SQuaJL}; the public crate also exports
\code{ClarkHash} aliases.

The contribution is an engineering package, not a claim that the underlying
mathematics is new. Johnson-Lindenstrauss projections
\cite{johnson1984extensions,achlioptas2003database,kane2014sparser}, feature
hashing \cite{weinberger2009feature}, and scalar quantization
\cite{gray1998quantization} are well studied. Clark Hash combines them into a
small deterministic codec for neural embedding storage:

\begin{enumerate}
  \item configure the input dimension, sketch dimension, bit width, sparsity,
        clip range, metric, and seed;
  \item encode each database vector independently;
  \item sketch each query in floating point; and
  \item score compressed vectors by an asymmetric inner product in sketch space.
\end{enumerate}

The implementation and benchmark harness are available at
\url{https://github.com/clark-labs-inc/clark-hash}.

\section{Related Work}

The Johnson-Lindenstrauss lemma shows that a finite set of vectors can be
embedded into a lower-dimensional Euclidean space while approximately
preserving pairwise distances \cite{johnson1984extensions}. Database-friendly
and sparse random projections reduce the arithmetic or randomness needed for
these transforms
\cite{achlioptas2003database,kane2014sparser,dasgupta2010sparse}. Feature
hashing applies signed hashing as a stateless dimensionality-reduction method
for high-dimensional sparse features \cite{weinberger2009feature}; related
sketching ideas also appear in data-stream algorithms such as CountSketch
\cite{charikar2002finding}.

Vector compression for retrieval often uses learned quantizers, including
product quantization and its variants \cite{jegou2011product}. These methods
can give better quality when a representative training set is available. Clark Hash
takes a different tradeoff. It gives up corpus-specific adaptation in exchange
for stateless online encoding and a small implementation.

Sentence embeddings are commonly evaluated by correlation with human similarity
judgments. Sentence-BERT popularized transformer-based sentence embeddings for
semantic textual similarity \cite{reimers2019sentencebert}, while MiniLM provides
small transformer backbones through self-attention distillation
\cite{wang2020minilm}. We evaluate on multilingual STS datasets surfaced through
MTEB \cite{muennighoff2022mteb}, including SemEval STS17
\cite{cer2017semeval} and STS22-style cross-lingual similarity data.

\section{Method}

\subsection{Sparse signed projection}

Let $x \in \R^d$ be a database embedding, $r \in \R^d$ a query embedding, $m$
the sketch dimension, $s$ the number of hash updates per input coordinate, $b$
the number of bits per quantized sketch coordinate, $c$ the symmetric clipping
range, and $\sigma_j(i) \in \{-1,+1\}$ the sign hash for input coordinate $i$
and repetition $j$. Bucket hashes $h_j(i)$ map coordinates to
$\{1,\ldots,m\}$. Clark Hash uses the sparse random matrix
\begin{equation}
R_{k,i} = \frac{1}{\sqrt{s}} \sum_{j=1}^{s} \sigma_j(i) \mathbf{1}[h_j(i)=k].
\end{equation}
Equivalently, the projected coordinate is
\begin{equation}
y_k = \sum_{i=1}^{d} \sum_{j=1}^{s}
      \mathbf{1}[h_j(i)=k] \sigma_j(i) \frac{x_i}{\sqrt{s}}.
\end{equation}
The projection is data-oblivious and deterministic given the seed. For fixed
vectors $u$ and $v$, the signed-hash estimator is centered:
\begin{equation}
\E[\inner{Ru}{Rv}] = \inner{u}{v}.
\end{equation}
Its variance depends on sketch dimension, sparsity, and collisions. Increasing
$m$ reduces projection noise at the cost of storage; increasing $s$ usually
reduces sparse-projection noise at the cost of encode CPU.

\subsection{Direction normalization and rescaling}

For cosine search, Clark Hash sketches the unit direction:
\begin{equation}
u = \frac{x}{\norm{x}}.
\end{equation}
The raw sketch is $Ru$. For a unit vector, each sketch coordinate has scale
roughly $1/\sqrt{m}$, so Clark Hash rescales by $\sqrt{m}$ before quantization:
\begin{equation}
z = \sqrt{m}\,Ru.
\end{equation}
This keeps the coordinate scale stable enough for a fixed clipping range such as
$[-3,3]$.

\subsection{Fixed scalar quantization}

Let $L=2^b-1$. Each scaled coordinate is clipped and uniformly quantized:
\begin{align}
z'_k &= \operatorname{clip}(z_k,-c,c),\\
q_k &= \operatorname{round}\!\left(L\frac{z'_k+c}{2c}\right).
\end{align}
The database-side dequantizer is
\begin{equation}
\hat{z}_k = \frac{2c q_k}{L} - c.
\end{equation}
The quantization step is $\Delta = 2c/L$. Without clipping, the scalar
quantization error per coordinate is bounded by
\begin{equation}
|\hat{z}_k - z_k| \leq \frac{\Delta}{2}.
\end{equation}
Clipping adds the residual $z_k-\operatorname{clip}(z_k,-c,c)$. Users can set
the clip range to trade clipping rate against quantization resolution.

\subsection{Asymmetric scoring}

Queries remain in floating point. For query $r$, Clark Hash computes
\begin{equation}
v = \frac{r}{\norm{r}}, \qquad a = \sqrt{m}\,Rv.
\end{equation}
The database vector stores the quantized sketch $\hat{z}$. The asymmetric
cosine estimate is
\begin{equation}
\widehat{\cos}(r,x) = \frac{1}{m} \sum_{k=1}^{m} a_k \hat{z}_k.
\end{equation}
Without quantization, the estimator maps back to cosine scale:
\begin{equation}
\frac{1}{m}\inner{\sqrt{m}Rv}{\sqrt{m}Ru}
= \inner{Rv}{Ru}
\approx \inner{v}{u}
= \cos(r,x).
\end{equation}
The floating-point query sketch avoids quantizing both sides of the score.
A quantization-only perturbation bound is
\begin{equation}
|\operatorname{score\_error}|
\leq \frac{1}{m}\sum_{k=1}^{m}|a_k|\frac{\Delta}{2},
\end{equation}
plus the analogous clipping term.

\subsection{Dot-product mode}

Cosine mode stores only the normalized direction sketch. Dot-product mode adds a
two-byte log-norm side channel:
\begin{align}
\ell &= \operatorname{clip}(\log_2\norm{x},\ell_{\min},\ell_{\max}),\\
n &= \operatorname{round}\!\left(
65535 \frac{\ell-\ell_{\min}}{\ell_{\max}-\ell_{\min}}
\right).
\end{align}
On decode,
\begin{equation}
\widehat{\norm{x}} = 2^{\hat{\ell}},
\end{equation}
and the final score is
\begin{equation}
\widehat{\operatorname{dot}}(r,x)
= \widehat{\cos}(r,x)\norm{r}\widehat{\norm{x}}.
\end{equation}

\section{Implementation}

The reference implementation is a Rust crate. After configuration, the codec is
stateless. It stores the sketch parameters, seed, quantizer levels, and metric.
Bucket locations and signs are generated from the seed, input dimension index,
and repetition index using a SplitMix64-style integer hash. Encoded database
vectors store the bit-packed coordinates and, in dot-product mode, the optional
two-byte norm channel. A query sketch stores a floating-point sketch and the
original query norm.

The computational cost to encode one vector is $O(ds+m)$: $ds$ sparse updates,
followed by quantization of $m$ sketch coordinates. Scoring one compressed
database vector costs $O(m)$. The included \code{FlatIndex} scans compressed
vectors exactly in sketch space. It is a reference tool, not an approximate
nearest-neighbor index.

For cosine mode, the storage cost per vector is
\begin{equation}
\left\lceil\frac{mb}{8}\right\rceil
\end{equation}
bytes. Dot-product mode adds two bytes. Table~\ref{tab:storage} gives the
default 384-dimensional sentence-embedding profile.

\begin{table}[t]
\centering
\caption{Default storage profile for 384-dimensional sentence embeddings.}
\label{tab:storage}
\begin{tabular}{lrr}
\toprule
Representation & Bytes per vector & Ratio vs. dense \code{f32} \\
\midrule
Dense \code{f32}, 384 dimensions & 1536 & 1.00000 \\
Clark Hash, $m=96$, $b=4$, cosine mode & 48 & 0.03125 \\
\bottomrule
\end{tabular}
\end{table}

\section{Evaluation}

\subsection{Setup}

We evaluated the default cosine profile with $m=96$, $b=4$, $s=4$, clip range
$c=3$, and seed 12345. This stores each 384-dimensional sentence embedding in
48 bytes, which is 32x smaller than dense \code{f32}. The benchmark downloads
multilingual sentence-similarity corpora from Hugging Face, embeds each unique
sentence once, encodes all sentence embeddings with Clark Hash, and compares
dense cosine scores with sketch scores.

The benchmark covers 9,304 labeled sentence pairs, 17,000 unique sentences, and
29 multilingual subsets from \code{mteb/sts17-crosslingual-sts} and
\code{mteb/sts22-crosslingual-sts}. Metrics are macro-averaged across language
subsets. For each subset we report dense cosine correlation with human labels,
Clark Hash correlation with human labels, Spearman loss relative to dense
cosine, and Pearson correlation between Clark Hash scores and dense cosine
scores.

We used two MiniLM-family encoders. The first, \code{all-MiniLM-L6-v2}, is
mostly an English model, so it is a stress test on cross-lingual data. The
second, \code{paraphrase-multilingual-MiniLM-L12-v2}, is multilingual and gives
a more direct view of the quantization loss on these corpora.

\subsection{Results}

\begin{table}[t]
\centering
\footnotesize
\setlength{\tabcolsep}{5pt}
\caption{Multilingual sentence-similarity results. Dense and sketch are macro
Spearman correlations with human labels. Loss is sketch minus dense.
Sketch/dense is macro Pearson correlation between sketch scores and dense cosine
scores.}
\label{tab:sts-results}
\begin{tabular}{llrrrrr}
\toprule
Model & Dataset & Pairs & Dense & Sketch & Loss & Sketch/dense \\
\midrule
MiniLM-L6 & STS17 & 5346 & 0.3644 & 0.2719 & -0.0926 & 0.7242 \\
MiniLM-L6 & STS22 & 3958 & 0.4168 & 0.2876 & -0.1292 & 0.8531 \\
MultiMiniLM-L12 & STS17 & 5346 & 0.8144 & 0.7460 & -0.0684 & 0.9099 \\
MultiMiniLM-L12 & STS22 & 3958 & 0.2973 & 0.2472 & -0.0501 & 0.9460 \\
\bottomrule
\end{tabular}

\vspace{0.35em}
\begin{minipage}{0.94\linewidth}
\footnotesize
MiniLM-L6 is all-MiniLM-L6-v2. MultiMiniLM-L12 is
paraphrase-multilingual-MiniLM-L12-v2. STS17 is
mteb/sts17-crosslingual-sts. STS22 is mteb/sts22-crosslingual-sts.
\end{minipage}
\end{table}

The multilingual model has higher scores on STS17. Dense cosine reaches 0.8144 macro
Spearman, and the 48-byte sketch reaches 0.7460. The sketch tracks dense cosine
with 0.9099 macro Pearson correlation. On STS22, the multilingual encoder has a
lower dense score, but the sketch still tracks dense cosine with 0.9460 macro
Pearson correlation.

The \code{all-MiniLM-L6-v2} run shows that model fit matters. Dense cosine is
already weak on many cross-lingual subsets. Clark Hash adds loss, but the main
problem in that run is that the embedding model is not well matched to the data.

\begin{table}[t]
\centering
\small
\caption{Local benchmark stage timings from the recorded JSON reports. These
numbers are sanity checks, not cross-machine performance claims.}
\label{tab:timing}
\begin{tabular}{lrrr}
\toprule
Model & Embedding seconds & Quantize seconds & Score seconds \\
\midrule
MiniLM-L6 & 119.23 & 0.0253 & 0.0049 \\
MultiMiniLM-L12 & 208.30 & 0.0253 & 0.0046 \\
\bottomrule
\end{tabular}
\end{table}

In this benchmark, embedding takes most of the runtime. Quantization and scoring are
small parts of the local run. The report does not include hardware details, so
these numbers should not be read as general performance claims.

\section{Discussion and Limitations}

Clark Hash may fit cases where embeddings arrive online and users want a compact
representation without fitting codebooks or calibration tables. This is also its
main limitation. Learned quantizers can adapt to a corpus and may get better
quality at the same byte budget. Clark Hash should be viewed as a simple storage
codec and sketch scoring method, not a replacement for product quantization or
graph-based approximate nearest-neighbor indexes.

The current benchmark measures score preservation and correlation with human
labels on sentence-similarity corpora. It does not measure retrieval recall in
large production corpora, adversarial streams, or hybrid indexes. No fixed
sketch dimension can preserve every future pair in an unbounded stream. Users
should tune $m$, $b$, $s$, and $c$ for their embedding model and quality target.

\section{Conclusion}

Clark Hash combines sparse signed random projection, fixed scalar quantization,
and asymmetric sketch scoring into a stateless codec for neural embeddings. The
default sentence-embedding profile stores 384-dimensional vectors in 48 bytes
and can encode each database vector on its own. On multilingual STS data, the
48-byte sketches keep a large part of the dense-score behavior when the
embedding model fits the data. The main point is simple: compact storage,
deterministic encoding, no fitting stage, and a small Rust implementation.

\section*{Availability}

Source code, benchmark harnesses, and JSON benchmark reports are available at
\url{https://github.com/clark-labs-inc/clark-hash}.

\bibliographystyle{plain}
\bibliography{references}

\end{document}